\documentclass[runningheads]{llncs}
\usepackage{graphicx}
\usepackage{amsmath,amssymb} 
\usepackage{color}

\usepackage[]{subfigure}   
\usepackage{multirow}

\usepackage[linesnumbered]{algorithm2e}

\usepackage{booktabs}
\usepackage{tabularx}
\usepackage{url}

\DeclareMathOperator*{\argmin}{arg\,min}

\newcommand{\blue}{\textcolor[rgb]{0.0,0.0,1.0}}

\begin{document}
\pagestyle{headings}
\mainmatter

\def\ACCV20SubNumber{196}  

\title{Meta-Learning with Context-Agnostic Initialisations} 

\titlerunning{Meta-Learning with Context-Agnostic Initialisations}
\authorrunning{Perrett \emph{et al.}}

\author{Toby Perrett, Alessandro Masullo, Tilo Burghardt, \\Majid Mirmehdi, Dima Damen}
\institute{Department of Computer Science, University of Bristol, UK\\\texttt{Toby.Perrett@bristol.ac.uk}}

\maketitle

\begin{abstract}
Meta-learning approaches have addressed few-shot problems by finding initialisations suited for fine-tuning to target tasks. Often there are additional properties within training data (which we refer to as context), not relevant to the target task, which act as a distractor to meta-learning, particularly when the target task contains examples from a novel context not seen during training. 

We address this oversight by incorporating a context-adversarial component into the meta-learning process. This produces an initialisation which is both context-agnostic and task-generalised. We evaluate our approach on three commonly used meta-learning algorithms and four case studies. We demonstrate our context-agnostic meta-learning improves results in each case. First, we report few-shot character classification on the Omniglot dataset, using alphabets as context. An average improvement of 4.3\% is observed across methods and tasks when classifying characters from an unseen alphabet. Second, we perform few-shot classification on Mini-ImageNet, obtaining context from the label hierarchy, with an average improvement of 2.8\%.  Third, we perform few-shot classification on CUB, with annotation metadata as context, and demonstrate an average improvement of 1.9\%. Fourth, we evaluate on a dataset for personalised energy expenditure predictions from video, using participant knowledge as context. We demonstrate that context-agnostic meta-learning decreases the average mean square error by 30\%.

\end{abstract}

\section{Introduction}

\begin{figure}[t]
\centering
\subfigure[\emph{Character-based} split. \label{fig:split_character}]{\includegraphics[scale=0.19,trim={0 0 30 0}]{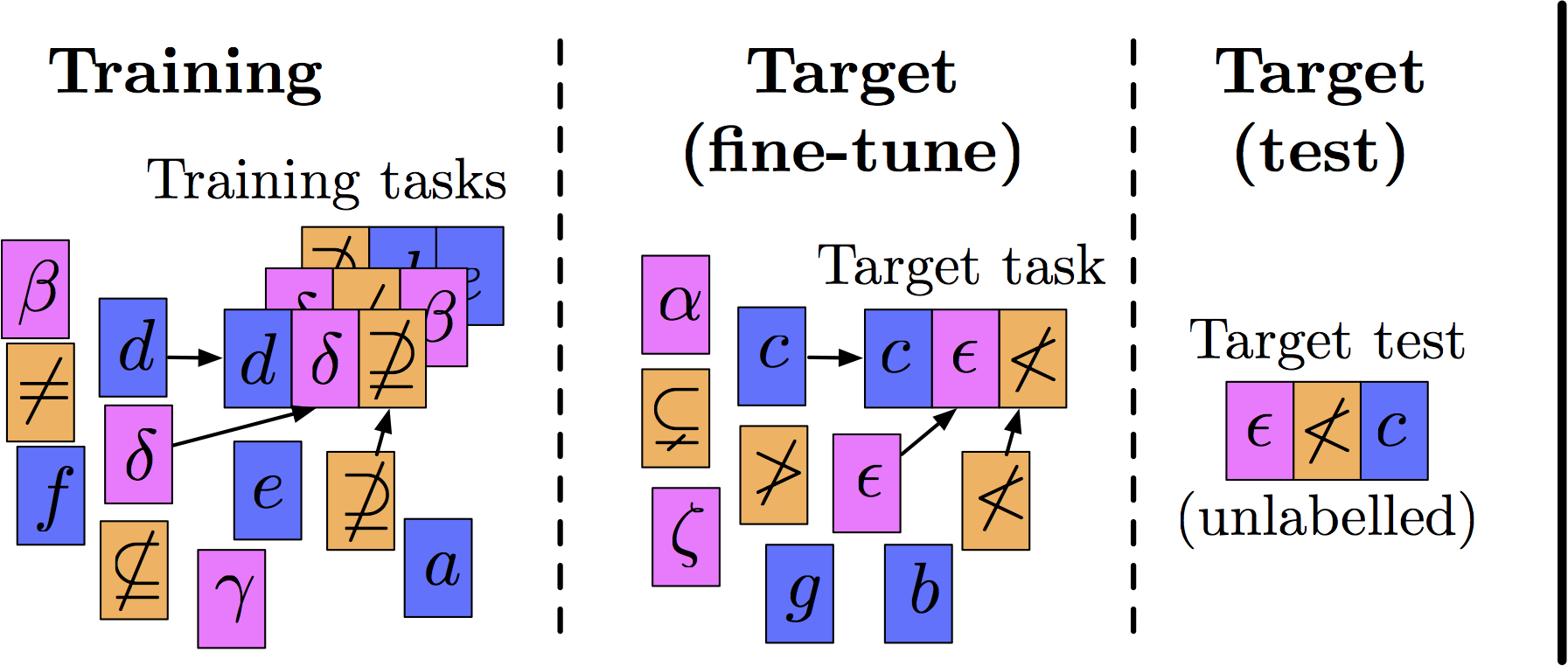}} 
\subfigure[\emph{Alphabet-based} split. \label{fig:split_alphabet}]{\includegraphics[scale=0.19]{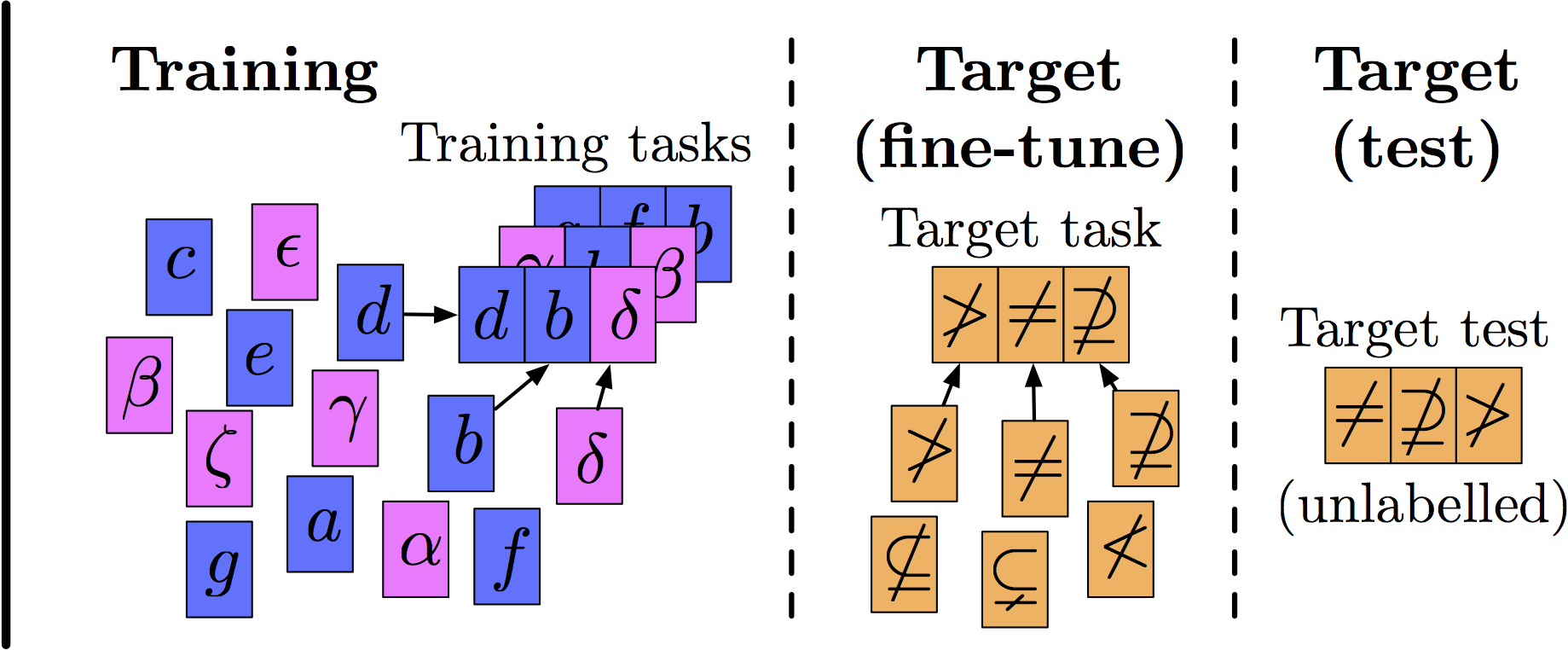}}  
\caption{Visualisation of how context (e.g. alphabets, shown as different colours) can contribute to train/target splits.  In commonly-used split (a), a classifier could overfit on context with no ill effects. If there is novel context, as in (b),
this will prove problematic.  In this paper, we show how context-agnostic meta-learning can benefit performance on few-shot target tasks without shared context.}
\label{fig:split}
\end{figure}

Current deep neural networks require significant quantities of data to train for a new task. 
When only limited labelled data is available, meta-learning approaches train a network initialisation on other \emph{source} tasks, so it is suitable for fine-tuning to new few-shot \emph{target} tasks~\cite{Finn2017}.  
Often, training data samples have additional properties, which we collectively refer to as \emph{context}, readily available through metadata.
We give as an example the \textit{alphabet} in a few-shot character recognition task (Fig. \ref{fig:split}).
This is distinct from multi-label problems as we pursue invariance to the context (i.e. alphabet), so as to generalise to unseen contexts in fine-tuning, rather than predicting its label.

In this work, we focus on problems where the target task is not only novel but does not have the same context as tasks seen during training. 
This is a difficult problem for meta-learners, as they can overfit on context knowledge to generate an initialisation, which affects the 
suitability for fine-tuning for tasks with novel contexts.
Prior works on meta-learning have not sought to exploit context, even when readily available~\cite{Finn2017,Rusu2019,Sun,Antoniou2018,Finn2018,Sun2019,Nichol,Bertinetto2018,Snell2017,Vinyals2016,Ren2018,Requeima2019,Tseng2020}.
We propose a meta-learning framework to tackle both task-generalisation and context-agnostic objectives, jointly. As with standard meta-learning, we aim for trained weights that are suitable for few-shot fine-tuning to target.
Note that concepts of \textit{context} and \textit{domain} might be incorrectly confused. Domains are typically different datasets with a significant gap, whereas context is one or more distractor signals within one dataset (e.g. font or writer for character classification), and can be either discrete or continuous.

\begin{figure}[t]
\centering
\subfigure[Randomly sample a task from all available training tasks.\label{fig:meta1}]{\includegraphics[scale=0.21]{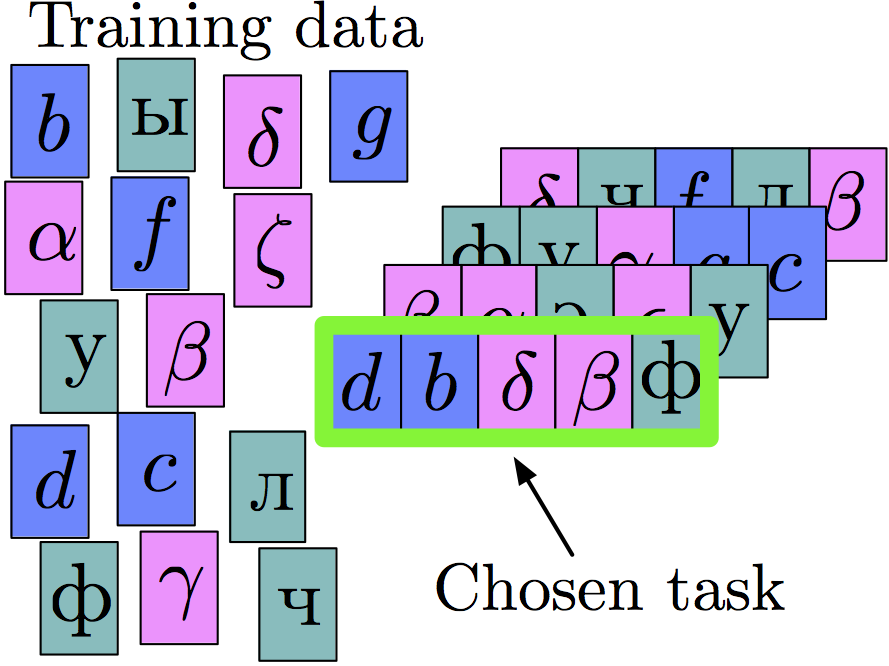}} \hspace{1mm}
\subfigure[Two copies are taken of the primary network weights.\label{fig:meta2}]{\includegraphics[scale=0.21]{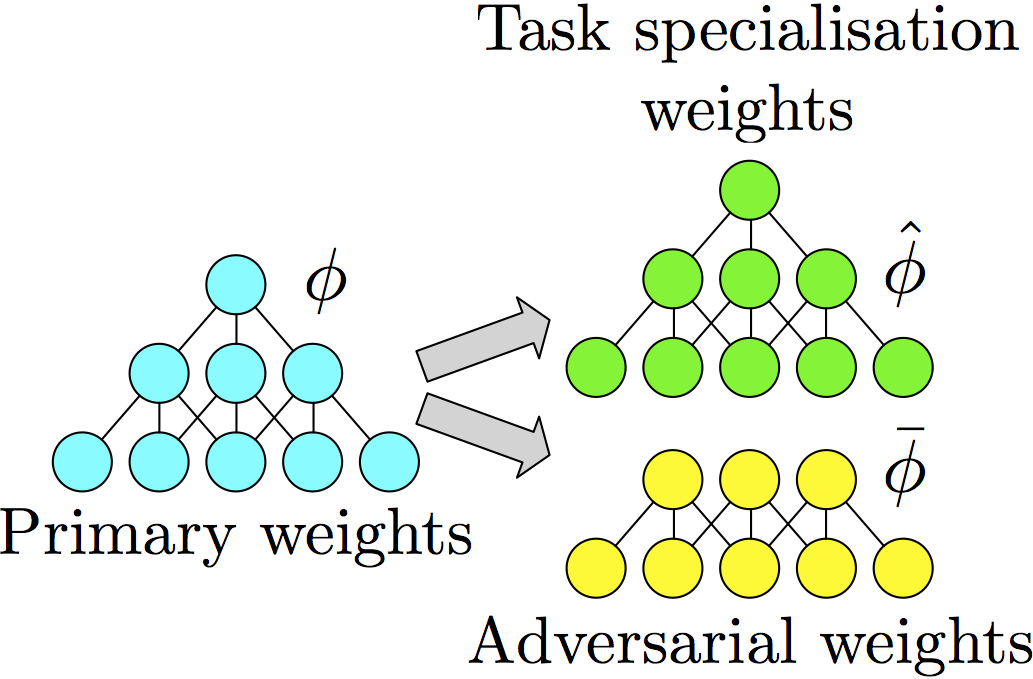}} \hspace{1mm}
\subfigure[$k$ rounds of optimisation on the chosen task, without context knowledge, to update $\hat{\phi}$.\label{fig:meta3}]{\includegraphics[scale=0.21,trim={-10 0 0 0}]{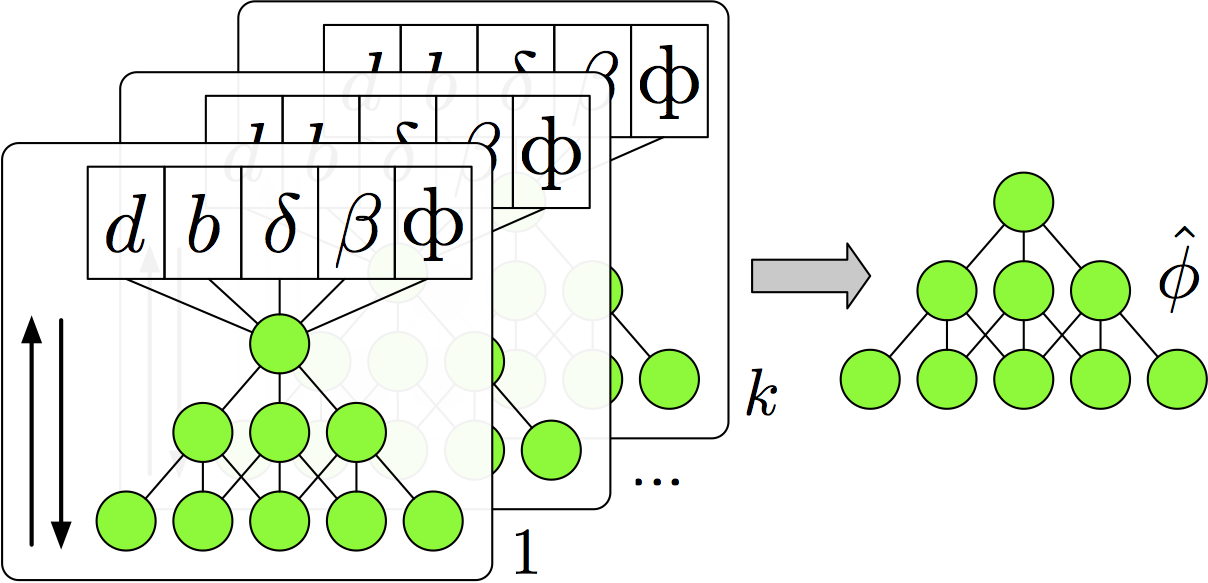}}\\
\subfigure[$l$ rounds of context-adversarial optimisation, passing the gradients though a gradient reversal layer to update $\bar \phi$.\label{fig:meta4}]{\includegraphics[scale=0.21]{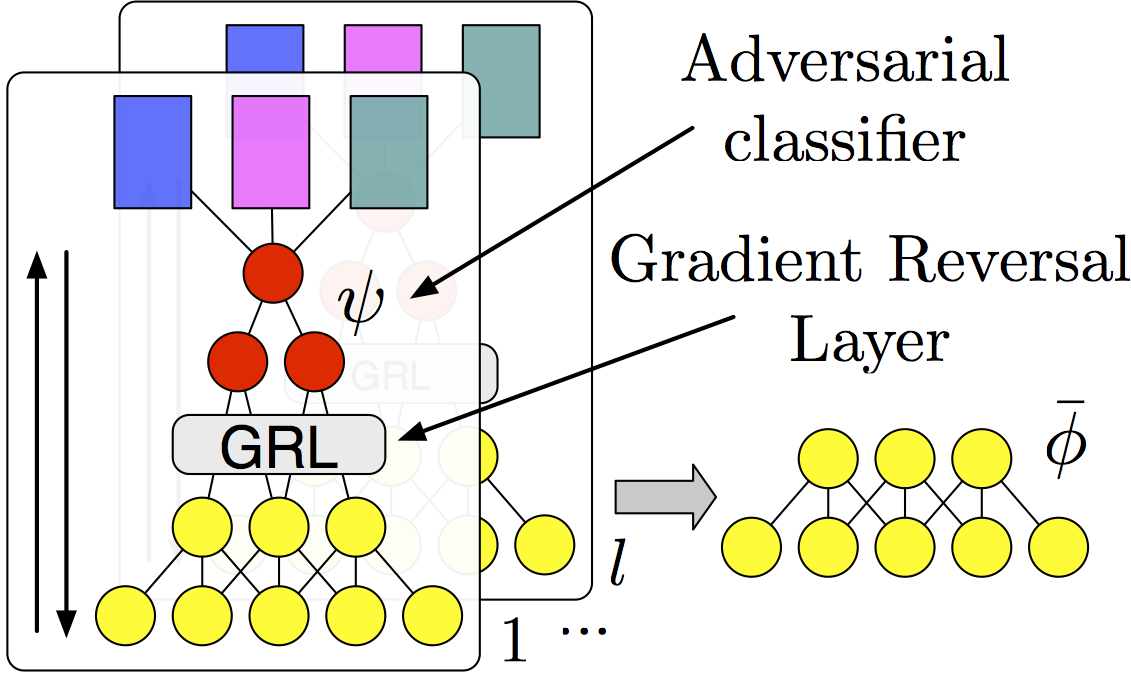}} \hspace{1mm}
\subfigure[Update primary weights from task-specific and context-agnostic optimisations. \label{fig:meta5}]{\includegraphics[scale=0.21]{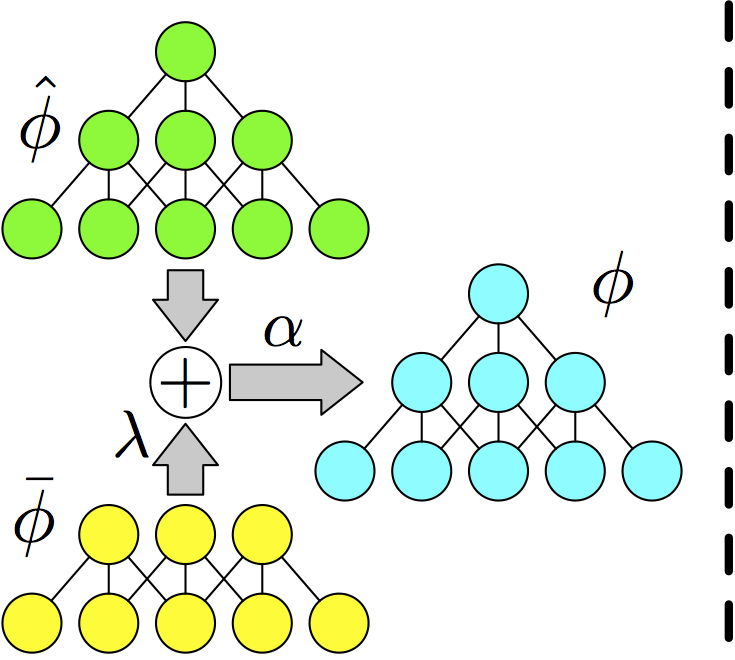}} \hspace{1mm}
\subfigure[After meta-learning,  the primary network can be fine-tuned for a new few-shot target task that might not share context with the training set.  \label{fig:meta6}]{\includegraphics[scale=0.21, trim={0 0 0 0}]{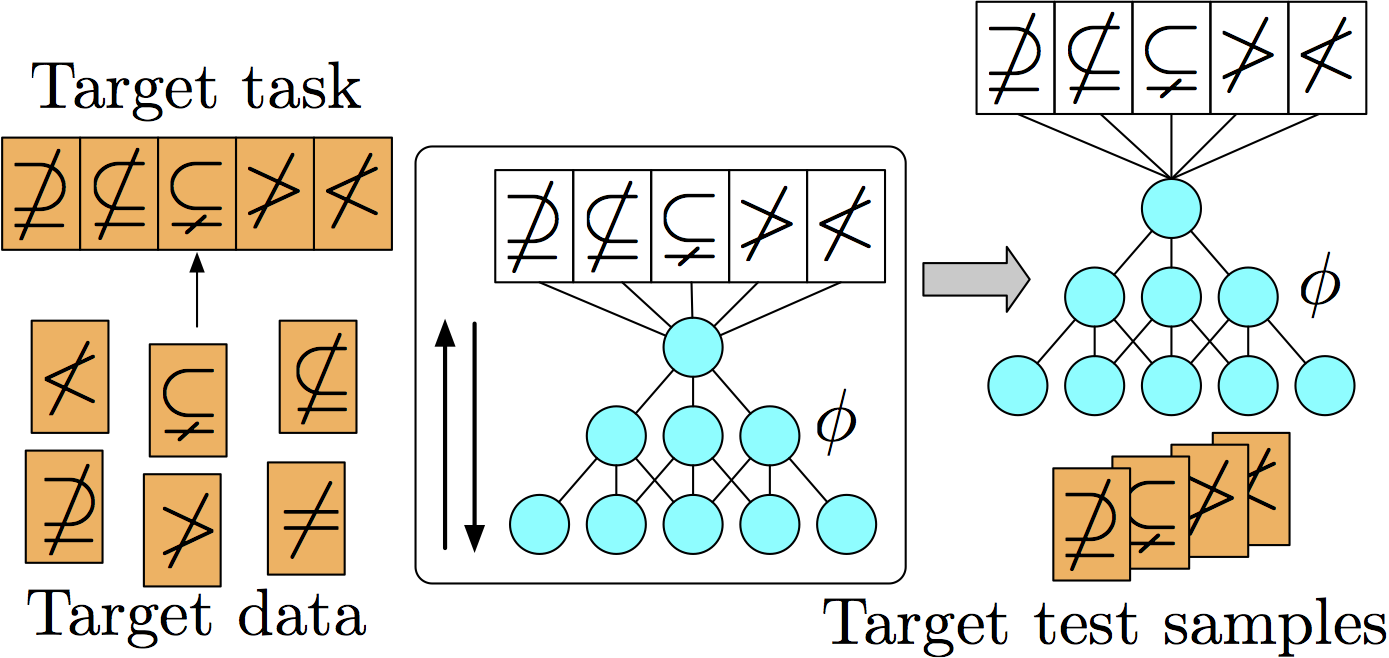}}
\caption{{A visualisation of the proposed context-agnostic meta-learning approach through a character classification example (context shown as character colours)} using an alphabet-based split (Fig.~\ref{fig:split_alphabet}). The method is detailed in Algorithm~\ref{alg:meta}, where (a) to (e) corresponds to one outer loop iteration, which is repeated on random training tasks.  (f) shows fine-tuning to target.}
\label{fig:meta}
\end{figure}

Figure~\ref{fig:meta} presents an overview of the proposed framework, illustrated on the application of character classification.
We assume that both task labels (e.g. character classification) and context labels (e.g. alphabet)  are available for the training data. At each iteration of meta-learning, we randomly pick a task~(Fig.~\ref{fig:meta1}), and optimise the model's weights for both task-generalisation~(Fig.~\ref{fig:meta3}) and context-agnosticism~(Fig.~\ref{fig:meta4}) objectives. This is achieved through keeping two copies of the model's weights (Fig.~\ref{fig:meta2}), one for each objective, and then updating the primary weights with a mixture of both results (Fig.~\ref{fig:meta5}). These learnt weights  are not only task-generalisable but importantly have been trained in an adversarial manner on context labels.

To demonstrate the generality of our framework, and the opportunities in considering context, we show that it is applicable to three commonly used few-shot meta-learning algorithms \cite{Finn2017,Antoniou2018,Nichol}, and
{test our context-agnostic meta-learning framework on four diverse problems, showing clear improvements compared to prior work and baselines.} The first problem (Sec~\ref{sec:omniglot}) is Omniglot character classification \cite{Lake2015}.  
We show that when using an alphabet-based split, our approach improves over non context-aware meta-learning approaches by 4.3\%. 
The second (Sec~\ref{sec:miniimagenet}) is Mini-ImageNet \cite{Vinyals2016} few-shot classification, where image classification is the task, and broader class group labels are the context. An improvement of 2.8\% is observed when utilising our approach. 
The third (Sec~\ref{sec:CUB}) is few-shot classification CUB \cite{WahCUB_200_2011}, where the primary colour of each bird (taken from annotations in metadata) is the context.  An improvement of 1.9\% is found in this case.
The fourth (Sec~\ref{sec:calorie}) is predicting energy expenditure of people performing daily activities from video~\cite{Tao}. For this problem, we consider calorie prediction as the task, and the identities as the context. We show that our approach drops the Mean Square Error (MSE) from 2.0 to 1.4.

\section{Related Work} \label{sec:related}

\noindent \textbf{Few-shot Learning:} Existing few-shot methods belong to one of three categories: generative approaches \cite{Zhang2018,Dwivedi2019}, embedding-based meta-learners \cite{Snell2017,Vinyals2016,Ren2018} and adaptation-based meta-learners \cite{Finn2017,Rusu2019,Sun,Antoniou2018,Finn2018,Sun2019,Nichol,Bertinetto2018,Requeima2019,Tseng2020}.
Adaptation-based meta-learners
produce initial models which can be fine-tuned quickly to unseen tasks, using limited labelled data.  One widely-used method is Model Agnostic Meta-Learning (MAML)~\cite{Finn2017}, where repeated specialisation on tasks drawn from the training set encourages the ability to adapt to new tasks with little data.  Later variations on this approach include promoting training stability~\cite{Antoniou2018} and improving training speed and performance on more realistic problems with deeper architectures~\cite{Nichol}.  Some works have learned alternative training curricula~\cite{Sun} or modified the task specialisation~\cite{Rusu2019,Bertinetto2018}. Others have learned alternative fine-tuning mechanisms \cite{Requeima2019,Tseng2020} or pseudo-random labels~\cite{Sun2019} to help with adaptation to unseen tasks.  
These adaptation-based meta-learners contrast with embedding-based meta-learners,
which find a space where the few-shot task can be embedded. A classifier is then constructed in this space, e.g. by comparing distances of target samples to seen source samples \cite{Vinyals2016}.

None of the above works have exploited context available from metadata of the training data.
Further, they have been evaluated on datasets where additional context knowledge is not available \cite{Oreshkin2018,Dwivedi2019}, where context is shared between the training and target split \cite{Lake2015,Vinyals2016} or combinations of the above \cite{triantafillou2019metadataset,Tseng2020}.
We select adaptation-based meta-learning as the most suitable candidate for few-shot tasks with context.  This is because there is likely to be insufficient target data for generative approaches, and target samples from a novel context are unlikely to embed well in the space constructed by embedding-based meta-learners.

\noindent \textbf{Domain Adaptation/Generalisation:} \hspace{6pt} 
Different from domains, contexts are additional labels present within the same dataset, can be continuous and one sample could be associated with multiple contexts. However, methods that attempt domain adaptation and generalisation are relevant for achieving context-agnostic learning.
Domain adaptation techniques aim to align source and target data.  
Some works use domain statistics to apply transformations to the feature space~\cite{Busto2017}, minimise alignment errors~\cite{Haeusser2017}, generate synthetic target data \cite{Hoffman2018,Huang2018} or learn from multiple domains concurrently~\cite{Rebuffi2017,Perrett2019,Li2019}. Adversarial domain classifiers have also been used to adapt a single \cite{Ganin2015,Zhang2019,Kang2018} and multiple \cite{Ros2019} source domains to a target domain.
The disadvantage of all these approaches  
is that sufficient target data is required, making them unsuitable for few-shot learning.
Domain generalisation works find representations agnostic to the dataset a sample is from.  Approaches include regularisation \cite{Balaji2018}, episodic training \cite{Li2019a,Dou2019a} and adversarial learning \cite{Li2018}.
In this paper, we build on adversarial training, 
as in~\cite{Ganin2015,Zhang2019,Kang2018,Ros2019,Li2018} for context-agnostic few-shot learning.

\section{Proposed Method}
We start Section~\ref{sec:problemFormulation} by formulating the problem, and explaining how it differs from commonly-tackled meta-learning problems.  In Section~\ref{sec:methodProposed}, we detail our proposal to introduce context-agnostic training during meta-learning. 

\subsection{Problem Formulation}
\label{sec:problemFormulation}
\noindent \textbf{Commonalities to other meta-learning approaches:} The input to our method is labelled training data for a number of tasks, as well as limited (i.e. few-shot) labelled data for target tasks. Adaptation-based meta-learning is distinct from other learning approaches in that the trained model is not directly used for inference. Instead, it is optimised for fine-tuning to a target task.
These approaches have two stages: (1) the meta-learning stage - generalisable weights across tasks are learnt, suitable for fine-tuning, and
(2) the fine-tuning to target stage - initialisation weights from the meta-learning stage are updated given a limited amount of labelled data from the target task.
This fine-tuned model is then used for inference on test data on the target task.
Throughout this section, we will focus on stage~(1), i.e. the meta-learning stage, as this is where our contribution lies.

\noindent \textbf{Our novelty:} We consider problems where the unseen target task does not share context labels with the training data.  We assume each training sample has both a task label and a context label.  The context labels are purely auxiliary - they are not the prediction target of the main network. 
We utilise context labels to achieve context-agnostic meta-learning using tasks drawn from the training set and argue that incorporating context-agnosticism provides better generalisation.  This is particularly important when the set of context labels in the training data is small, increasing the potential discrepancy between tasks.

\subsection{Context-Agnostic Meta-Learning}
\label{sec:methodProposed}
{Our contribution is applicable to adaptation-based meta-learning algorithms which are trained in an episodic manner.  This means they use an inner update loop to fine-tune the network weights on a single task, and an outer update loop which incorporates changes made by the inner loop into a set of primary network weights \cite{Finn2017,Rusu2019,Antoniou2018,Finn2018,Nichol}.
To recap, none of these algorithms exploit context knowledge, and although they  differ in the way they specialise to a single task in the inner loop, they all share a common objective:}
\begin{equation} \label{eq:obj1}
    \min_{\phi} \mathbb{E}_{\tau}\left[ L_{\tau} \left( U_{\tau}^k \left( \phi \right) \right) \right]  ,
\end{equation}
where $\phi$ are the network weights, $\tau$ is a randomly sampled task 
and $L_{\tau}$ is the loss for this task.  $U_{\tau}$ denotes an update which is applied $k$ times, using data from task $\tau$.
{Algorithm~\ref{alg:meta} shows (in black) the core of the method employed by \cite{Finn2017,Antoniou2018,Nichol}, including the inner and outer loop structure common to this class of meta-learning technique.
They differ in the way they calculate and backpropogate $\nabla L_{\tau}$ in the inner specialisation loop (where different order gradients are applied, and various other training tricks are used).
This step appears in Algorithm \ref{alg:meta} L7-10 and Fig. \ref{fig:meta3}.  However, they can all be modified to become context-agnostic in the same way - this is our main contribution (shown in blue in the algorithm), which we discuss next.}

To achieve context-agnostic meta-learning, we propose to train a context-adversarial network alongside the task-specialised network.  This provides a second objective to our meta-learning.
We update the meta-learning objective from Eq.~\ref{eq:obj1} to include this context-adversarial objective, to become
\begin{equation}
    \min_{\phi, \psi} \mathbb{E}_{\tau}\left[ L_{\tau} \left( U_{\tau}^k \left( \phi \right) \right) + \lambda L_C \left( U_C^l \left( \psi, \phi \right) \right)  \right] ,
    \label{eq:obj2}
\end{equation}
where $L_C$ is a context loss, given by an associated context network with weights $\psi$, which acts on the output of the network with weights $\phi$. $U_C \left(\psi, \phi \right)$ is the adversarial update which is performed $l$ times.
The relative contribution of $L_C$ is controlled by $\lambda$.  Because $L_C$ and $L_{\tau}$ both operate on $\phi$, they are linked and should be optimised jointly. 
Equation~\ref{eq:obj2} can thus be decomposed into two optimisations:
\begin{eqnarray} \label{eq:cond1}
    \phi\! &= &\!\argmin_{\phi} \left( L_{\tau} \left( U_{\tau}^k \left( \phi \right) \right) - \lambda L_C \left( U_C^l   \left(\psi, \phi \right) \right) \right) \\
\label{eq:cond2}
        \psi\! &= &\! \argmin_{\psi} \left( L_C \left( U_C^l \left(\psi, \phi \right)  \right) \right) .
\end{eqnarray}

\begin{algorithm}[t]
\SetAlgoLined{
 Initialise primary network with parameters $\phi$.\\
 \blue{Initialise adversarial network with parameters $\psi$.}\\
 \blue{Link primary and adversarial networks with GRL}\\
 \For{Iteration in outer loop}{
  Select random task $\tau$.\\
  Set $\hat \phi = \phi$ \blue{and $\bar \phi = \phi$}.\\
  \For{Iteration in inner specialisation loop}{
    Construct batch with samples from task $\tau$.\\
    Calculate $L_{\tau}$. \\
    Optimise $\hat \phi$ w.r.t. $L_{\tau}$.\\
  }
  \For{\blue{Iteration in inner adversarial loop}}{
    \blue{Construct batch with samples from training dataset.}\\
    \blue{Add context label noise with probability $\epsilon$.} \\
    \blue{Calculate $L_C$.} \\
    \blue{Optimise $\psi$ and $\bar \phi$ w.r.t. $L_C$}\\
  }
 Update \blue{$\phi \gets \phi + \alpha (\hat \phi - \phi + \lambda (\bar \phi - \phi))$.}
}
}
\caption{Context-agnostic meta-learning framework. Proposed additions which can be encapsulated by existing adaptation-based meta-learning approaches, such as \cite{Finn2017,Antoniou2018,Nichol}, are in blue. }
\label{alg:meta}
\end{algorithm}

We can observe the adversarial nature of $L_C$ in Eqs. \ref{eq:cond1} and \ref{eq:cond2}, 
where, {while} $\psi$ attempts to minimise $L_C$,
$\phi$ attempts to extract features which are context-agnostic (i.e. maximise $L_C$). 
To optimise, we proceed with two steps.  The first is to update the context predictor $\psi$ using the  gradient $ \nabla_{\psi} L_C(\psi, \phi)$.  This is {performed} $l$ times, which we write as
\begin{equation}\label{eq:grad1}
    U_C^l \left( \nabla_{\psi} L_C(\psi, \phi) \right).
\end{equation}
A higher $l$ means the adversarial network trains quicker, when balanced against $k$ to ensure $\psi$ and $\phi$ learn together in an efficient manner.
The second step is to update the primary network with weights $\phi$ with the gradient
\begin{equation}\label{eq:grad2}
     \nabla_{\phi} L_{\tau} \left( U_{\tau}^k(\phi) \right) - \lambda  \nabla_{\phi} L_C \left( U_C^l  (\psi, \phi) \right).
\end{equation}
The first term corresponds to the contribution of the task-specific inner loop.  The method in \cite{Nichol} reduces this quantity to $\left(\phi - U_{\tau}^k(\phi) \right) / \alpha$, where $\alpha$ is the learning rate.
$\lambda$~is a weighting factor for the contribution from the adversarial classifier, which can analogously be reduced to $\lambda \left(\phi - U_C^l  (\psi, \phi) \right) / \alpha$.  It can be incorporated by backpropagating the loss from $\psi$ through a gradient reversal layer (GRL) to~$\phi$.
As well as performing Eqs.~\ref{eq:grad1} and \ref{eq:grad2}, we also perform each iteration of the $l$ adversarial updates $U_C$ with respect to $\psi$ and $\phi$ concurrently.

In practice, the process above can be simplified by taking two copies of the primary weights at the start of the process as shown in Algorithm~\ref{alg:meta}, which matches the illustration in Fig.~\ref{fig:meta}. At each outer iteration, we first choose a task (Algorithm~\ref{alg:meta} L5) and make two copies of the primary weights $\phi$ (L6): $\hat \phi$ (weights used for the task-specialisation inner loop) and $\bar \phi$ (weights used for the context-adversarial inner loop).  The task specialisation loop is then run on~$\hat \phi$~(L7-10).  Next, the adversarial loop is run on $\bar \phi$ and $\psi$ (L12-17).  The primary weights~$\phi$ are updated using weighted contributions from task-specialisation ($\hat \phi$) and context-generalisation ($\bar \phi$)~(L18).
Note that using two separate copies of the weights ensures that the task-specialisation inner loop is as similar as possible to the one fine-tuned for the target task.

The optimiser state and weights for the adversarial network with weights $\psi$ are persistent between outer loop iterations so $\psi$ can learn context as training progresses.  This contrasts with the optimisers acting on the $\hat \phi$ and $\bar \phi$, which are reset every outer loop iteration for the next randomly selected task to encourage the initialisation to be suitable for fast adaptation to a novel task.

Following standard meta-learning approaches, the weight initialisations $\phi$ can be fine-tuned to an unseen target task. 
After fine-tuning on the few-shot labelled data from target tasks, this updated model can be used for inference on unlabelled data from these target tasks (see Fig. \ref{fig:meta6}).
No context labels are required for the target, as the model is trained to be context-agnostic. Our method is thus suitable for fine-tuning to the target task when new context is encountered, as well as when contexts overlap.

Next, we explore four problems for evaluation. Recall that our approach assumes both task and context labels are available during training. In all our cases studies, we select datasets where context is available, or can be discovered, from the metadata.

\section{Case Study 1: Character Classification}\label{sec:omniglot} 

\noindent \textbf{Problem Definition.} Our first case study is few-shot image classification benchmark - Omniglot~\cite{Lake2015}.  
We consider the task as character classification and the context as which alphabet a character is from.
We follow the standard setup introduced in \cite{Vinyals2016}, which consists of 1- and 5-shot learning on sets of 5 and 20 characters (5- or 20-way) from 50 alphabets.  However, we make one major and important change.
Recall, we have suggested that existing meta-learning techniques are not designed to handle context within the training set, or context-discrepancy between training and target.
The protocol from \cite{Vinyals2016} uses a \emph{character}-based split, where an alphabet can contribute characters to \textit{both} train and target tasks (Fig. \ref{fig:split_character}).  
Instead, we eliminate this overlap by ensuring that the characters are from different alphabets, i.e. an \emph{alphabet}-based split (Fig.~\ref{fig:split_alphabet}).

\noindent \textbf{Evaluation and Baselines.}
{We evaluate the proposed context-agnostic framework using three meta-learners: MAML++ ~\cite{Antoniou2018}, MAML~\cite{Finn2017} and REPTILE~\cite{Nichol}.  Note that other adaptation-based meta-learning methods could also be used by substituting in their specific inner-specialisation loops \cite{Rusu2019,Finn2018}.  Unmodified versions are used as baselines, and are compared against versions which are modified with our proposed context agnostic (CA) component.} We accordingly refer to our modified algorithms as CA-MAML++, CA-MAML and CA-REPTILE.
We report results without transduction, that is batch normalisation statistics are not calculated from the entire target set in advance of individual sample classification. This is more representative of a practical application.  As in~\cite{Vinyals2016}, the metric is top-1 character classification accuracy.  
We run experiments on the full dataset, and also on a reduced number of alphabets. 
With 5 alphabets, for example, characters from 4 alphabets are used for training, and a few-shot task is chosen from the 5th alphabet only.
As the number of alphabets in training decreases, a larger context gap would be expected between training and target.
We report averages over 10 random train/target splits, and keep these splits consistent between experiments on the same number of alphabets.

\noindent \textbf{Implementation Details.}
The widely-used architecture, optimiser and hyperparameters introduced in \cite{Vinyals2016}, are used. We implement the adversarial context predictor in the proposed context-agnostic methods as a single layer which takes the penultimate features layer (256D) as input with a cross-entropy loss  applied to the output, predicting the alphabet.  Context label randomisation is used in the adversarial classifier, where 20\% of the context labels are changed. This stops the context adversarial loss tending to zero too quickly (similar to label smoothing~\cite{Salimans2016}).  {We use $l=3$ (Eq. \ref{eq:obj2}) for all Omniglot experiments. The context-agnostic component increases the training time by 20\% for all methods.}

\begin{figure}[t]
    \centering
    \subfigure[Averaged over the 1- and 5-shot, 5- and 20-way tasks, showing the effect of the number of unique context labels (i.e. alphabets). \label{fig:diff_ab}]{\includegraphics[scale=0.43, trim={0 0 0 0}]{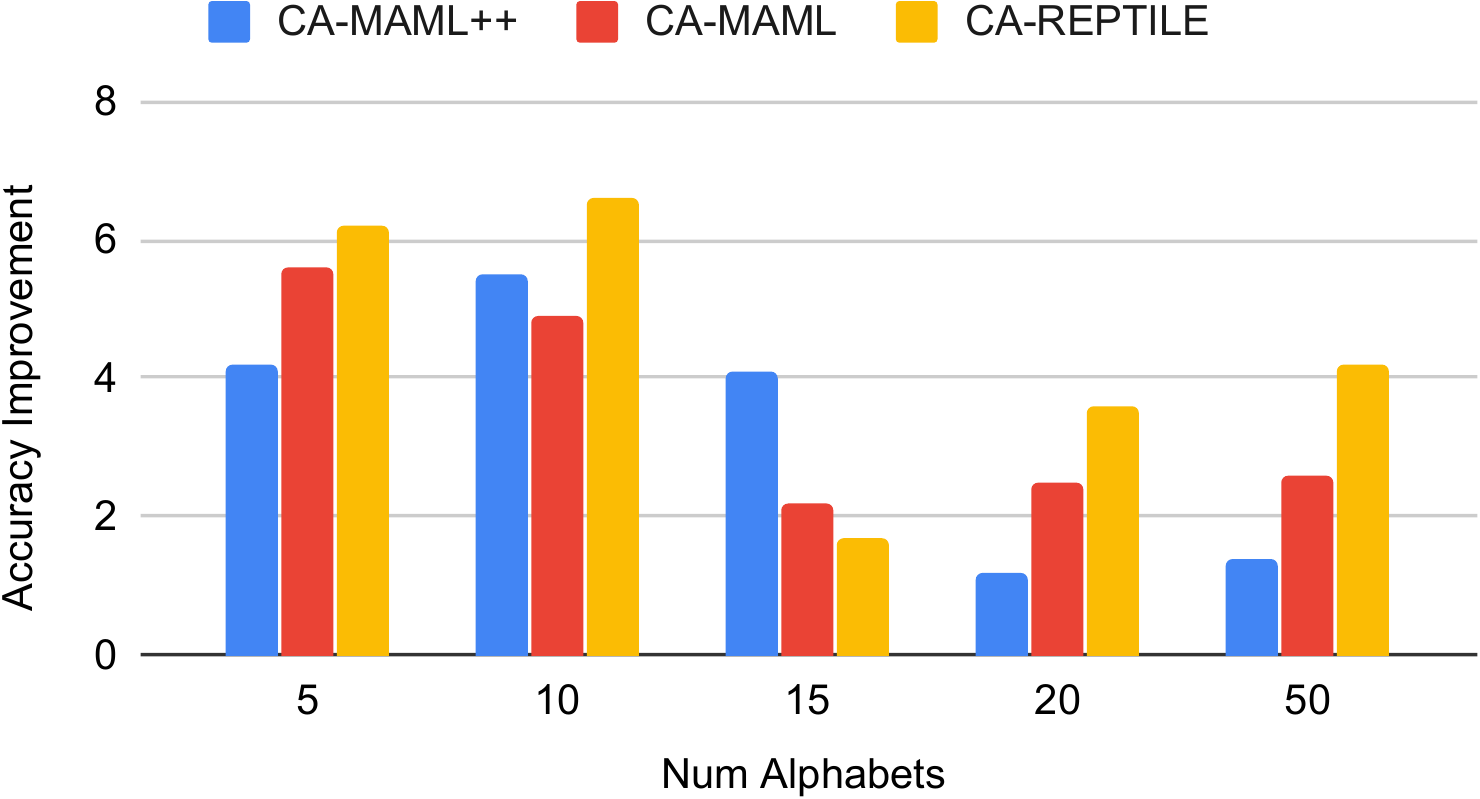}} \hspace{1mm}
    \subfigure[Averaged over number of alphabets (5, 10, 15, 20 and 50), showing how each task is affected.\label{fig:diff_task}]{\includegraphics[scale=0.43, trim={0 -1 0 0}]{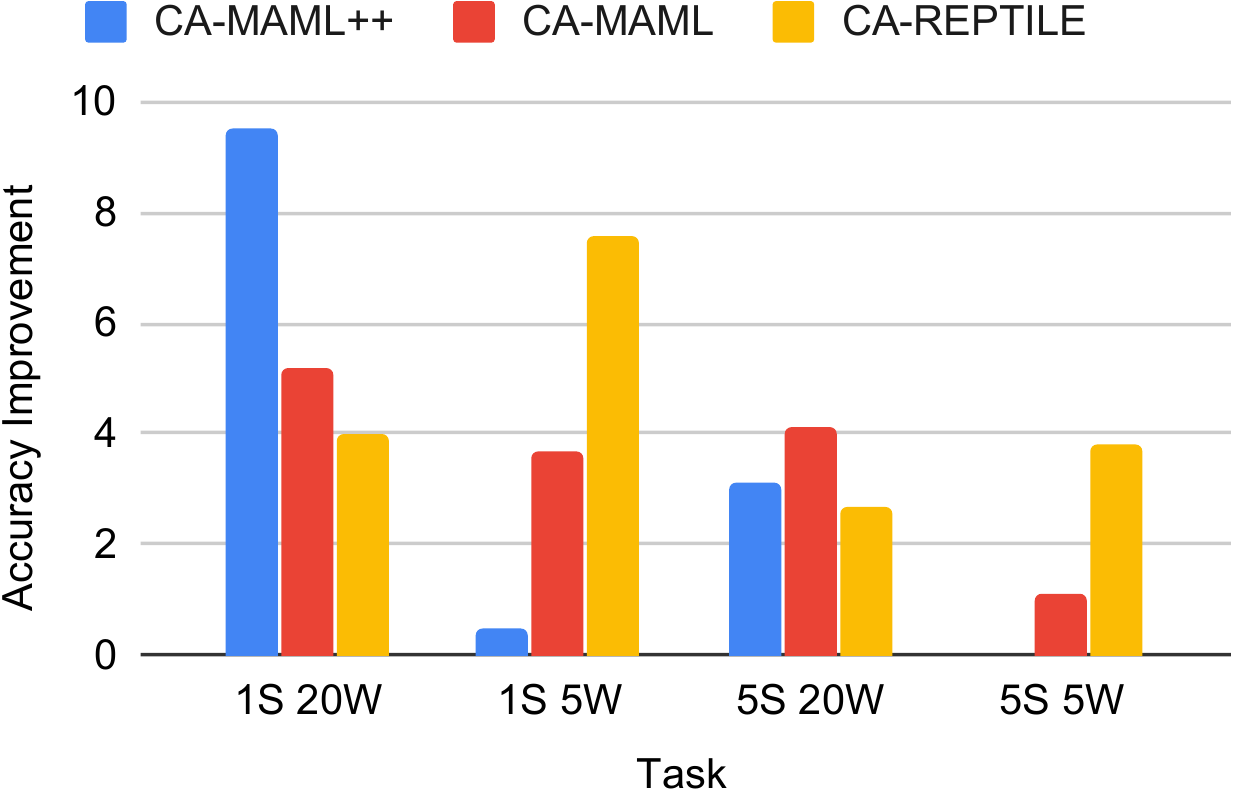}} 
    \caption{{Accuracy improvements given by our context-agnostic (CA-) versions of \cite{Antoniou2018,Finn2017,Nichol} using the alphabet-based split (shown in Fig. \ref{fig:split_alphabet}).}}
    \label{fig:diff}
\end{figure}

\noindent {\bf Results.}
{Table \ref{tab:alphabets_all} shows the results of the proposed framework applied to \cite{Antoniou2018,Finn2017,Nichol} on 5-50 alphabets, using the alphabet-based split shown in Fig.~\ref{fig:split_alphabet}. We report results per method, to show our proposed context-agnostic component improves on average across all methods, tasks and numbers of alphabets. 85\% of individual method/task/alphabet combinations show an improvement, with a further 10\% being comparable (within 1\% accuracy).  Overall, the proposed framework gives an average performance increase of 4.3\%.  This improvement is most pronounced for smaller numbers of alphabets (e.g. average improvements of $>=$6.2\%, 4.9\% and 4.2\%  for 5 and 10 alphabets for \cite{Nichol,Finn2017,Antoniou2018} respectively). 
This trend is shown in Fig. \ref{fig:diff_ab}, and} supports our earlier hypothesis that the inclusion of a context-agnostic component is most beneficial when the context overlap between the train and target data is smaller.
Fig. \ref{fig:diff_task} shows the improvement for each XS YW task, averaged over the number of alphabets.  Larger improvements are observed for all methods on the 1-shot versions of 5- and 20-way tasks, with \cite{Nichol} improving the most on 1S 5W and \cite{Finn2017,Antoniou2018} improving the most on 1S 20W.

\begin{table}[t]
\centering
\caption{Character classification accuracy on Omniglot, using an alphabet-based split, with the number of training alphabets varied between 5 and 50. XS YW indicates X-shot fine-tuning at a Y-way classification tasks.  Base methods are compared against context-agnostic (CA) versions.}
\resizebox{1\textwidth}{!}{%
\begin{tabular}{llrrrrr}
			\toprule
			 & & \multicolumn{5}{c}{Number of Alphabets}  \\ 
			 \cmidrule(){3-7}
			Task \hspace{5mm}                    & Method \hspace{10mm}          & \hspace{25pt}5         & \hspace{15pt}10        & \hspace{15pt}15        & \hspace{15pt}20        & \hspace{15pt}50        \\ \midrule
			\multirow{6}{*}{1S 20W} & MAML++ \cite{Antoniou2018}    & 58.7 & 57.2 & 64.7 & \bf{85.6} & 89.6 \\
			                        & CA-MAML++                     & \bf{72.3} & \bf{67.6} & \bf{82.4} & 84.8 & \bf{90.9} \\ \cmidrule(){2-7}
			                        & MAML \cite{Finn2017}          & 61.4 & 78.2 & 81.5 & 83.7 & 87.5 \\
			                        & CA-MAML                       & \bf{69.8} & \bf{82.8} & \bf{82.1} & \bf{89.8} & \bf{93.8} \\ \cmidrule(){2-7}
			                        & REPTILE \cite{Nichol}         & 11.9    & 18.1     & 37.6    & 51.6    & 64.9      \\
			                        & CA-REPTILE                    & \bf{20.7}    & \bf{21.8}     & \bf{39.5}    & \bf{55.5}    & \bf{66.5} \\ \midrule	
			\multirow{6}{*}{1S 5W}  & MAML++ \cite{Antoniou2018}    & 97.4 & 96.2 & \bf{94.9} & 93.4 & 93.7 \\
			                        & CA-MAML++                     & \bf{98.1} & \bf{97.1} & 90.1 & \bf{95.8} & \bf{97.1} \\ \cmidrule(){2-7}
			                        & MAML \cite{Finn2017}          & 86.1 & 87.0 & \bf{96.1} & 94.4 & 90.5 \\
			                        & CA-MAML                       & \bf{94.5} & \bf{91.3} & 94.7 & \bf{96.0} & \bf{96.2} \\ \cmidrule(){2-7}
			                        & REPTILE \cite{Nichol}         & 52.2    & 68.8    & 79.4     & 75.5    & 77.5      \\
			                        & CA-REPTILE                    & \bf{62.2}    & \bf{76.9}    & \bf{83.4}     & \bf{83.2}    & \bf{85.5} \\ 
			                        \bottomrule
\end{tabular}
\hspace{0.05\linewidth}
\begin{tabular}{llrrrrr}
			\toprule
			 & & \multicolumn{5}{c}{Number of Alphabets}  \\ 
			 \cmidrule(){3-7}
			Task \hspace{5mm}                    & Method \hspace{10mm}          & \hspace{25pt}5         & \hspace{15pt}10        & \hspace{15pt}15        & \hspace{15pt}20        & \hspace{15pt}50        \\ \midrule
			\multirow{6}{*}{5S 20W} & MAML++ \cite{Antoniou2018}    & 81.0 & 84.1 & 92.4 & 93.5 & 95.8 \\
			                        & CA-MAML++                     & \bf{84.8} & \bf{90.8} & \bf{96.0} & \bf{94.5} & \bf{96.3} \\ \cmidrule(){2-7}
			                        & MAML \cite{Finn2017}          & 81.7 & 83.8 & 84.0 & 91.2 & \bf{89.0} \\
			                        & CA-MAML                       & \bf{86.0} & \bf{91.8} & \bf{92.9} & \bf{93.1} & 86.9 \\ \cmidrule(){2-7}
			                        & REPTILE \cite{Nichol}         & 58.4    & 68.1    & 76.7    & \bf{76.0}    & 78.0      \\
			                        & CA-REPTILE                    & \bf{61.1}    & \bf{73.7}     & \bf{78.3}    & 75.8    & \bf{81.6} \\ \midrule
			\multirow{6}{*}{5S 5W}  & MAML++ \cite{Antoniou2018}    & \bf{99.4} & \bf{99.3} & \bf{98.7} & 97.0 & 96.8 \\
			                        & CA-MAML++                     & 99.3 & 98.6 & 98.5 & \bf{99.4} & \bf{96.9} \\ \cmidrule(){2-7}
			                        & MAML \cite{Finn2017}          & 96.6 & 95.8 & 97.2 & 97.9 & 98.9 \\
			                        & CA-MAML                       & \bf{97.8} & \bf{98.5} & \bf{97.6} & \bf{98.6} & \bf{99.1} \\ \cmidrule(){2-7}
			                        & REPTILE \cite{Nichol}         & 85.2      & 85.6      & \bf{93.2} & 88.5      & 89.4      \\
			                        & CA-REPTILE                    & \bf{88.3} & \bf{94.4} & 92.4      & \bf{91.6} & \bf{92.9} \\ 
			\bottomrule
		\end{tabular}
		}
\label{tab:alphabets_all}
\end{table}

\begin{figure}[t]
    \centering
    \subfigure[Accuracy on  the training set after the inner loops.]{\includegraphics[width=0.49\columnwidth, trim={0 0 0 25}]{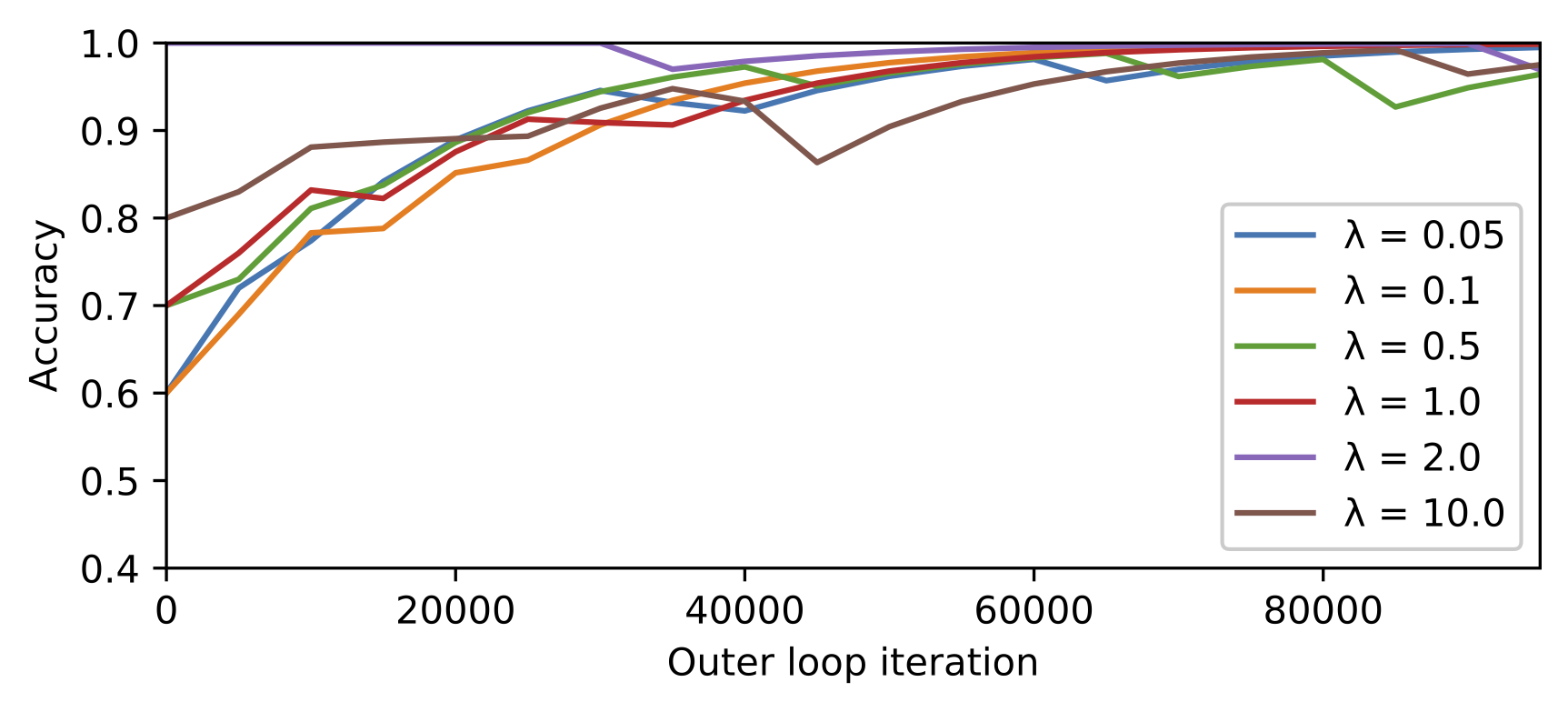}}
    \subfigure[Accuracy on the target set after fine-tuning to the target task.]{\includegraphics[width=0.49\columnwidth, trim={0 0 0 20}]{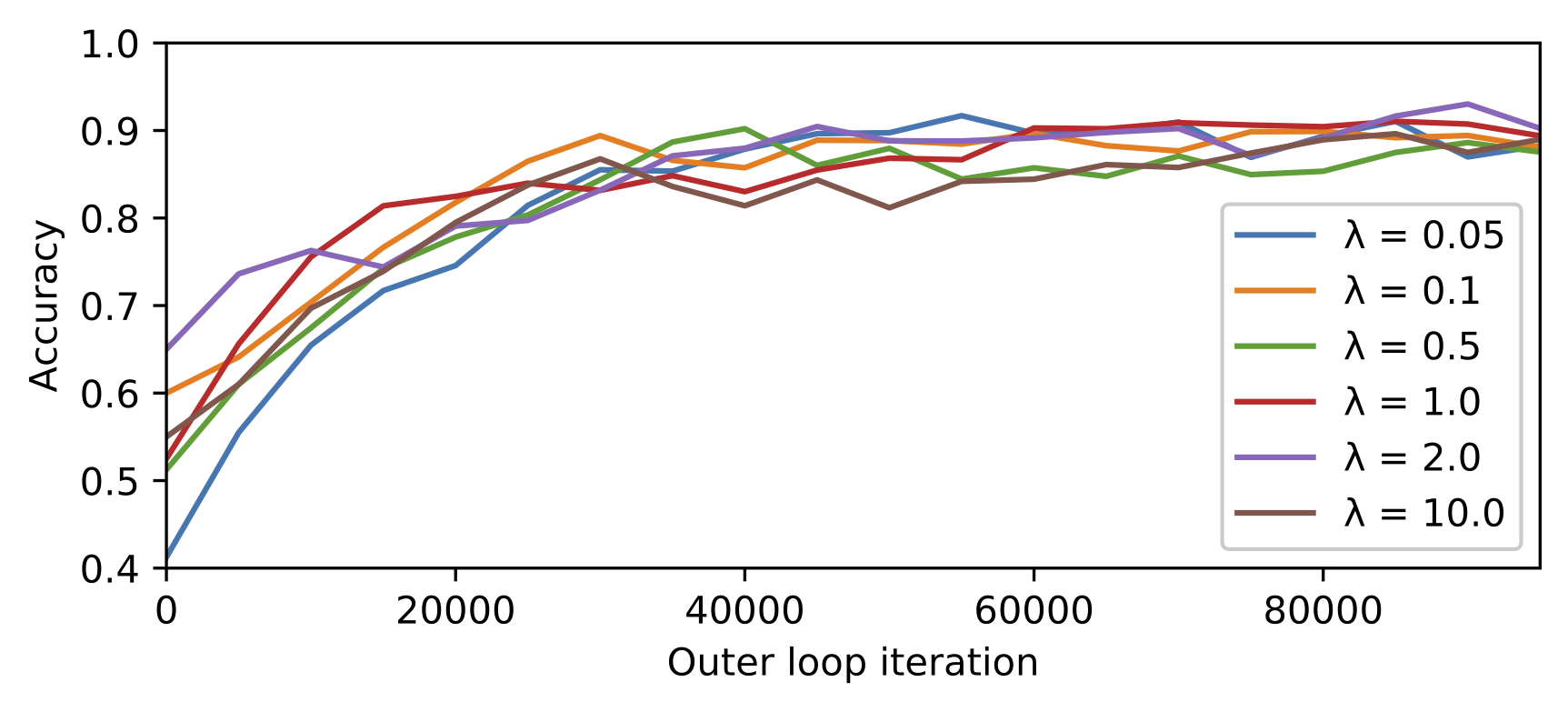}} 
    \caption{These plots show how the weighting ($\lambda$) of the context-adversarial component affects training and target performance during one run of the 5-shot/5-way 10 alphabet task using an alphabet-based split.}
    \label{fig:lambda}
\end{figure}

For the ablation studies, we use \cite{Nichol} as our base meta-learner as it is the least computationally expensive.  Based on preliminary studies, we believe the behaviour is consistent, and the conclusions stand, for the other methods.
{In the results above, we used $\lambda=1.0$ for the contribution of our adversarial component $\lambda$ (Eq.~\ref{eq:obj1}). Next,} we provide results on how varying $\lambda$ can affect the model's performance. 
For this, we use 5S 5W, 10 alphabet task.  Fig. \ref{fig:lambda} shows training progress with $\lambda = \{10.0, 2.0, 1.0, 0.5, 0.1\}$.  We can see that a high weighting ($\lambda = 10.0$) causes a drop in training accuracy around iteration 40K, as the optimisation prioritises becoming context-agnostic over the ability to specialise to a task. However, the figure shows reasonable robustness to the choice of $\lambda$.

\begin{figure}[t]
    \centering
    \subfigure[50 alphabets. \label{fig:ab50}]{\includegraphics[width=0.49\columnwidth, trim={0 0 0 0}]{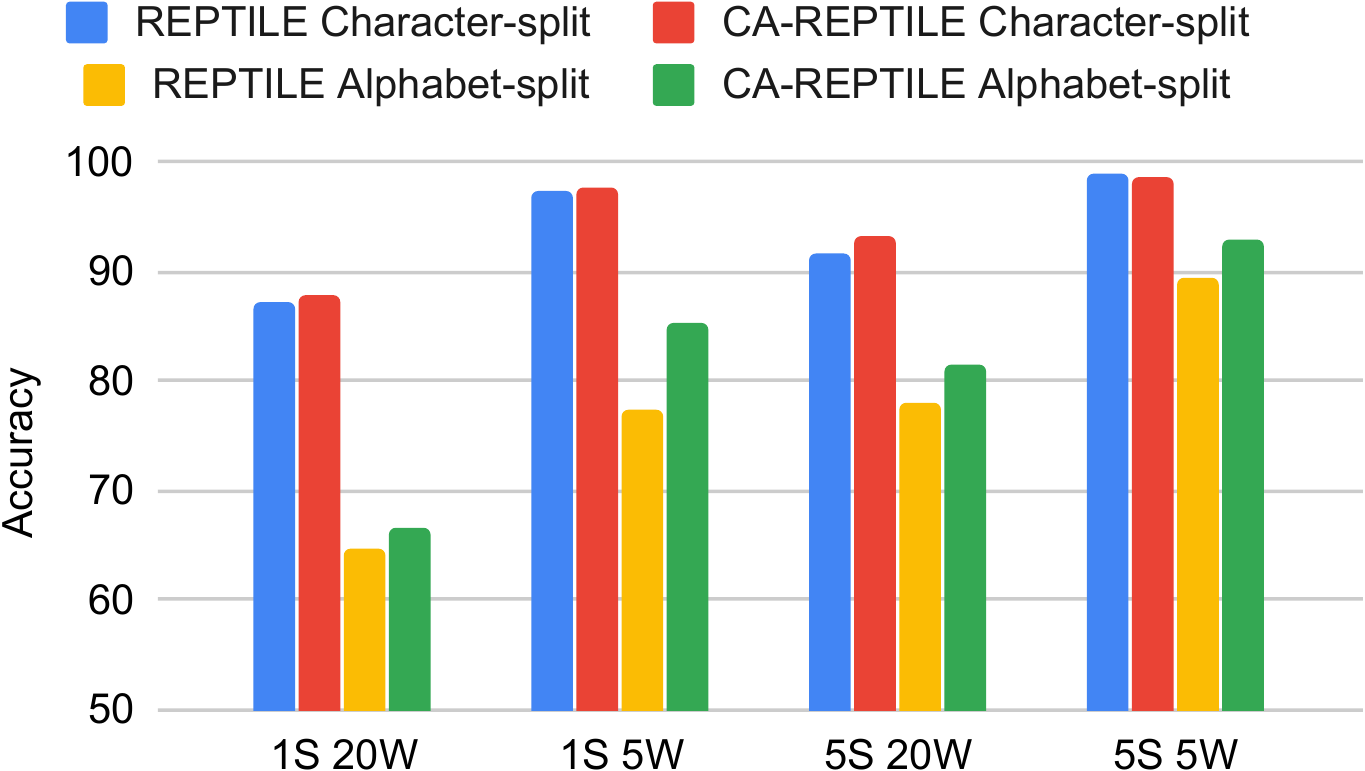}}
    \subfigure[10 alphabets.\label{fig:ab10}]{\includegraphics[width=0.49\columnwidth, trim={0 0 0 0}]{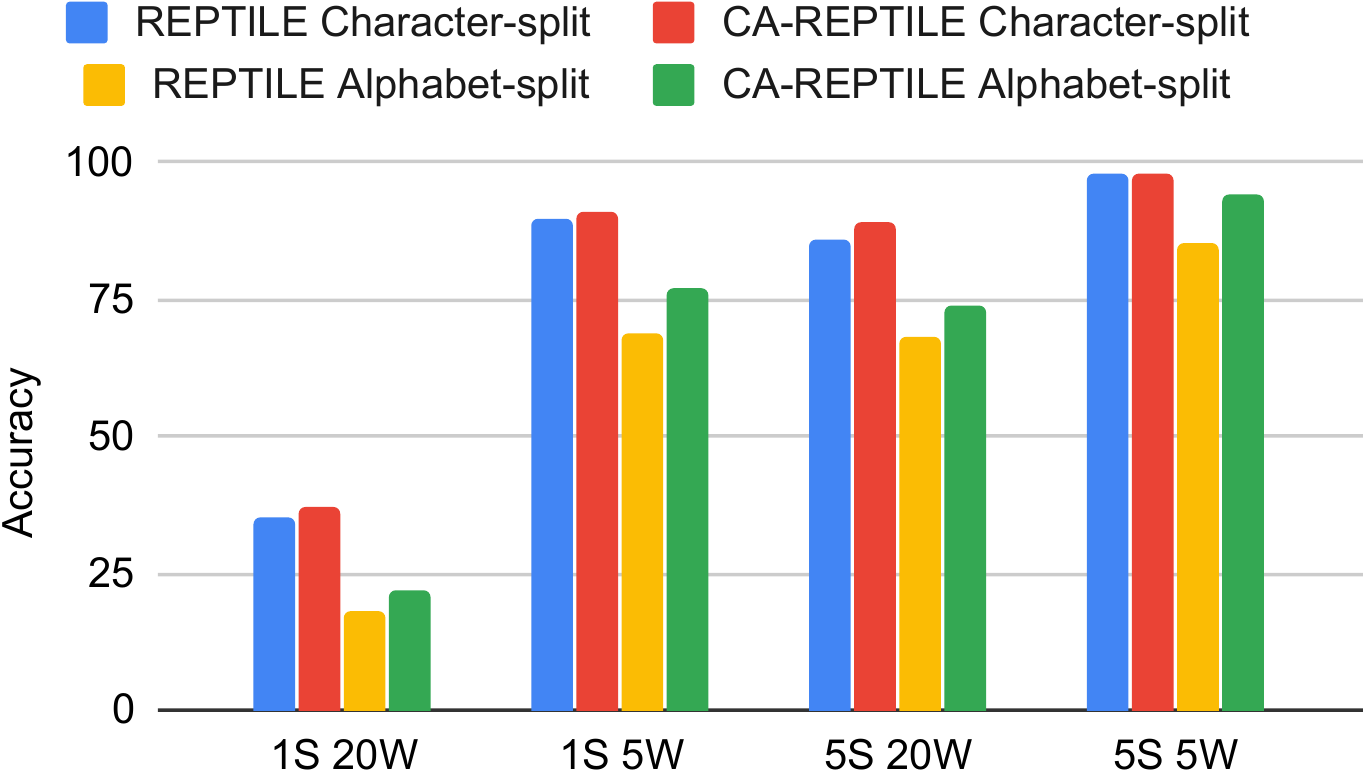}} 
    \caption{{Comparison of character-based and alphabet-based training/target splits using 50 and 10 alphabets.}}
    \label{fig:ab}
\end{figure}

Next, we investigate the differences between character-based and alphabet-based  training/target splits (visualised in Fig. \ref{fig:split}).  
Fig.~\ref{fig:ab} 
shows the effects of context-agnosticism when evaluating on character-based splits and alphabet-based splits.
Fig. \ref{fig:ab50} uses 50 alphabets for comparison, and Fig. \ref{fig:ab10} uses 10 alphabets. 
While both approaches are comparable on character-based splits (blue vs red), we show a clear improvement in using our context-agnostic meta-learning approach when tested on alphabet-based splits (yellow vs green). This is a sterner test due to the training and target sets being made up from data with different contexts. The context-agnostic version is significantly better for all cases and both alphabet sizes.

Finally, as previous approaches only evaluate on the easier character-based split for Omniglot, using all 50 alphabets, we provide comparative results to published works on this setup. We list reported results from \cite{Finn2017,Antoniou2018,Nichol} as well as our replications to ensure a direct comparison (the same codebase and splits can be used with and without the context-agnostic component). 
For this setup, we use the same data augmentation as~\cite{Finn2017,Antoniou2018,Nichol}.
Results are given in Table \ref{tab:omniglot_results_1},
which confirms that context-agnostic versions of the base methods achieve comparable performance, despite there being shared context between source and target.

In summary, this section presented experiments on the Omniglot character classification dataset. We show that, on average, our proposed context-agnostic approach gives performance improvements across all {methods and} tasks, particularly for smaller alphabet sizes, which introduce a bigger context gap between training and target.

\begin{table}[t]
\caption{Comparative results on Omniglot using the standard character-based split. *: results reported in cited papers.  Even though both training and target tasks share context, our  CA contribution maintains performance on this standard split. }
\centering 
\resizebox{0.53\textwidth}{!}{
\begin{tabular}{lrrrr}
\toprule
Method & \hspace{2mm}5S 5W & \hspace{2mm}1S 5W & \hspace{2mm}5S 20W & \hspace{2mm}1S 20W \\ \midrule
MAML++ \cite{Antoniou2018}*    & 99.9         & 99.4        & 99.3        & 97.7          \\ 
MAML++ \cite{Antoniou2018}    & 99.9         &  99.5       & 98.7        & 95.4          \\ 
CA-MAML++                      & 99.8         & 99.5        & 98.8        & 95.6          \\ \midrule
MAML \cite{Finn2017}*         & 99.8         & 98.6        & 98.9        & 95.8          \\ 
MAML \cite{Finn2017}         & 99.8          & 99.3        & 97.0        & 92.3          \\ 
CA-MAML         & 99.8         & 99.3        & 97.2        &  94.8         \\ \midrule
REPTILE \cite{Nichol}*           & 98.9           & 95.4           &  96.7           &  88.1           \\ 
REPTILE \cite{Nichol}            & 98.9           & 97.3           &  96.4           &  87.3           \\ 
CA-REPTILE                    & 98.6           & 97.6           &  95.9           &  87.8           \\
\bottomrule
\end{tabular}
}
\label{tab:omniglot_results_1}
\end{table}

\section{Case Study 2: General Image Classification}\label{sec:miniimagenet}

\noindent \textbf{Problem Definition.} Our second case study uses the few-shot image classification benchmark - Mini-ImageNet~\cite{Vinyals2016}.  We use the experimental setup introduced in \cite{Vinyals2016}, where the task is a 1- or 5-shot 5-way classification problem.
Similar to our previous case study, we aim for context labels, and a context-based split.
This dataset has no readily-available context labels, and there is a large overlap between the train and target splits (e.g. 3 breeds of dog in target, 12 in train). We address this by manually assigning 12 superclass labels, which we use as context. 
We then ensure that superclasses used for training and testing are distinct.

\noindent \textbf{Evaluation, Baselines and Implementation.}
Similar to Section \ref{sec:omniglot}, we evaluate using MAML++ ~\cite{Antoniou2018} and MAML~\cite{Finn2017}.  Unmodified versions are used as baselines, and are compared against versions which are modified with our proposed CA component.  Transduction is not used, and the metric is top-1 image classification accuracy.  The same architecture, hyperparameters etc. as in~\cite{Antoniou2018} are used.  We use $k=5$ (Eq. \ref{eq:obj1}) and $l=2$ (Eq. \ref{eq:obj2}). Results are given for the original Mini-ImageNet splits and our superclass-based splits with context labels.

\noindent \textbf{Results.}  Table \ref{tab:mi_results} shows the results on the original train/target split and the new splits with no shared context.  
Results show comparable performance for the original split, but importantly improved performance in the context-based split.
Our context-agnostic component improves over \cite{Finn2017} and \cite{Antoniou2018} by an average 3.3\% on the most difficult 1S 5W task.  An average 2.2\% improvement is also seen on the easier 5S 5W task. Similar to Omniglot, note that few shot classification on Mini-ImageNet is more challenging (by an average of 8.7\% across all methods) when there is no shared context between training and target data.

\begin{table}[t]
\caption{Results on Mini-ImageNet and CUB using the original splits which have shared context between train and target tasks, and the new context-based splits with no shared context between training and target tasks.}
\centering 
\resizebox{0.8\textwidth}{!}{%
\begin{tabular}{lrrrrrrrr}
\toprule
            & \multicolumn{4}{c}{Mini-ImageNet} & \multicolumn{4}{c}{CUB} \\
			 & \multicolumn{2}{c}{Original split} & \multicolumn{2}{c}{Context Split} & \multicolumn{2}{c}{Original split} & \multicolumn{2}{c}{Context Split}\\ 
			 \cmidrule(lr){2-3} \cmidrule(lr){4-5} \cmidrule(lr){6-7} \cmidrule(lr){8-9}
Method & \hspace{2mm}1S 5W & \hspace{2mm}5S 5W & \hspace{2mm}1S 5W & \hspace{2mm}5S 5W & \hspace{2mm}1S 5W & \hspace{2mm}5S 5W & \hspace{2mm}1S 5W & \hspace{2mm}5S 5W\\ \midrule
MAML++ \cite{Antoniou2018}    & \bf{52.0}        & \bf{68.1}      & 40.1         & 60.1       & \bf{38.7}        & 57.2      & 42.2         & 56.7            \\ 
CA-MAML++                     & 51.8        & \bf{68.1}      & \bf{44.4}         & \bf{61.5}       & 38.0        & \bf{58.4}      & \bf{43.3}         & \bf{57.9}            \\ \midrule
MAML \cite{Finn2017}          & \bf{48.3}        & \bf{64.3}      & 41.1         & 56.5       & 42.5        & \bf{56.1}      & 37.7         & 54.7            \\ 
CA-MAML                       & \bf{48.3}        & 64.2      & \bf{43.3}         & \bf{59.5}       & \bf{42.6}        & 55.9      & \bf{40.3}         & \bf{57.5}           \\
\bottomrule
\end{tabular}
}
\label{tab:mi_results}
\end{table}


\section{Case Study 3: Fine-Grained Bird Classification}\label{sec:CUB}

\noindent \textbf{Problem Definition.} For our third case study, we use the few-shot fine-grained bird classification benchmark CUB \cite{WahCUB_200_2011}.  CUB contains a large amount of metadata from human annotators.  For context labels, we have taken each bird's primary colour, but could have chosen a number of others e.g. bill shape. The CUB dataset has 200 classes, with 9 different primary colours.  We ensure splits are distinct with respect to this property.

\noindent \textbf{Evaluation, Baselines and Implementation.}
We use the same setup as for Mini-ImageNet (Section \ref{sec:miniimagenet}).

\noindent \textbf{Results.}  Table \ref{tab:mi_results} shows the results on the original train/target splits and the new splits with no shared context (i.e. no shared primary colour).  When there is less less shared context between train and target data, our context-agnostic component improves over \cite{Finn2017} and \cite{Antoniou2018} by an average of 1.9\% across all tasks, whilst performance is maintained on the original split.

\section{Case Study 4: Calorie Estimation from Video}\label{sec:calorie}

\noindent \textbf{Problem Definition.} In this fourth problem,  we use the dataset from \cite{Tao2018}, where the task is to {estimate energy expenditure for an input video sequence of an indiviual carrying out a variety of actions.}
Different from the first three case studies, this is a regression task, rather than a classification one, as calorie readings are continuous.
The target task is to estimate the calorimeter reading for seen, as well as unseen, actions. Importantly, the individual captured forms the context.  Alternative context labels could include, for example, age or Body Mass Index (BMI).
Our objective is thus to perform meta-learning to generalise across actions, as well as being individual-agnostic, for calorie prediction of a new individual.
We use silhouette footage and calorimeter readings from 10 participants performing a number of daily living tasks as derived from the {SPHERE Calorie dataset} of~\cite{Tao}.
Using a relatively small amount of data to fine-tune to target is appropriate because collecting data from individuals using a calorimeter is expensive and cumbersome.

\noindent \textbf{Evaluation and Baselines.} Ten-fold leave-one-person-out cross-validation is used for evaluation. We report results using MSE across all videos for each subject.  For fine-tuning to target, we use labelled calorie measurements from the first 32 seconds (i.e. the first 60 video samples, where each sample is 30 frames subsampled at 1fps) of the target subject.  Evaluation is then performed using the remaining data from the target subject, which is 28 minutes on average.
We compare the following methods, using cross-fold, leave-one-person-out validation:
\begin{itemize}
    \item Metabolic Equivalent (MET) from \cite{Tao}. This offers a baseline of calorie estimation through a look-up table of actions and their duration. This has been used as a baseline on this dataset previously.
    \item Method from Tao et al. \cite{Tao2018} that utilises IMU and depth information not used by our method.
    \item Pre-train - standard training process, trained on 9 subjects and tested on target subject without fine-tuning.
    \item Pre-train/fine-tune - standard training process on 9 subjects and fine-tuned on the target subject.
    \item REPTILE - meta-learning from~\cite{Nichol} on 9 subjects and fine-tuned on target.
    \item CA-REPTILE - our proposed context-agnostic meta-learning approach.
\end{itemize}
Note that we chose to use \cite{Nichol} as the baseline few-shot method because it is less computationally expensive (important when scaling up the few shot-problem to video) than \cite{Finn2017,Antoniou2018}, as discussed in Section \ref{sec:related}.

\noindent \textbf{Implementation Details.}
Images are resized to 224x224, and fed to a ResNet-18 architecture \cite{He2016}.  No previous works have addressed this individual-agnostic personalisation problem. Following~\cite{Tao}, it is believed that a window of 30s is required as input for energy expenditure prediction. We sample the data at 1fps and use the ResNet CNN's output from the penultimate layer as input to a Temporal Convolutional Network (TCN) \cite{Bai2018} for temporal reasoning. 
Our model is trained end-to-end using Adam \cite{Kingma2015} and contains 11.2M parameters.  {We use $k=10$ (Eq. \ref{eq:obj1}) and $l=1$ (Eq. \ref{eq:obj2}) for all Calorie experiments.  A lower value of $l$ is required than for Omniglot, as context information is easier for the adversarial network to learn (i.e. people are easier to distinguish than alphabets).}
MSE is used as the regression loss function.  
Augmentation during training consists of random crops and random rotations up to 30$^\circ$.
The same architecture is used for all baselines (except MET and \cite{Tao2018}), making results directly comparable.

\begin{table}[t]
\caption{MSE for all 10 participants on the Calorie dataset, using leave-one-out cross-validation.  A lower MSE indicates better results. Methods with only an average reported are results taken from the referenced publications.}
\centering 
\resizebox{0.8\textwidth}{!}{
\begin{tabular}{lrrrrrrrrrrr}
\toprule
{Method}     & {P1} & {P2} & {P3} & {P4} & {P5} & {P6} & {P7} & {P8} & {P9} & {P10} & {\hspace{2mm}Avg} \\ \midrule
MET Lookup \cite{Tao}       &  -       &  -       & -        & -       & -        & -        & -        & -        & -        & -         & 2.25         \\ 
Tao et al. \cite{Tao2018}  &  -       &  -       & -        & -       & -        & -        & -        & -        & -        & -         & 1.69         \\ 
Pre-train only        &  1.21       &  \bf{0.89}       & 0.88        & 1.86        & 1.24        & \bf{2.46}        & 7.50        & \bf{0.89}        & 1.25        & 3.11         & 2.13         \\ 
Pre-train/fine-tune & 0.58        & 1.64        & 0.75        &  0.53       & 1.13        & 4.26        & 5.83        & 1.29        & 1.41        & 3.53         & 2.10         \\ 
REPTILE  \cite{Nichol}  &  0.48 & 1.65 & 0.52 & 0.90 & 2.12 & 3.28 & 6.48 & 1.26 & \bf{0.83} & 2.58 & 2.01 \\ 
CA-REPTILE            & \bf{0.39}        & 1.11        & \bf{0.46}        & \bf{0.48}        & \bf{0.87}        & 2.68        & \bf{3.75}        & 1.07        & 0.87        & \bf{2.32}         & \bf{1.40}                  \\
\bottomrule
\end{tabular}
}
\label{tab:cal_main_results}
\end{table}

\noindent{\bf Results.}
Table \ref{tab:cal_main_results} compares the various methods.  
The context-agnostic meta-learning method obtains a 35\% reduction in MSE over the pre-training only, a 33\% reduction over the pre-train/fine-tune model, and a 30\% improvement over the non context-agnostic version.
For 3 out of 10 individuals, pre-training outperforms any fine-tuning. We believe this is due to these participants performing actions at the start of the sequence in a different manner to those later. However, our context-agnostic approach offers the best fine-tuned results.

\begin{figure}[t]
\centering
\subfigure{\includegraphics[width=\textwidth, trim={-250 0 -40 0}]{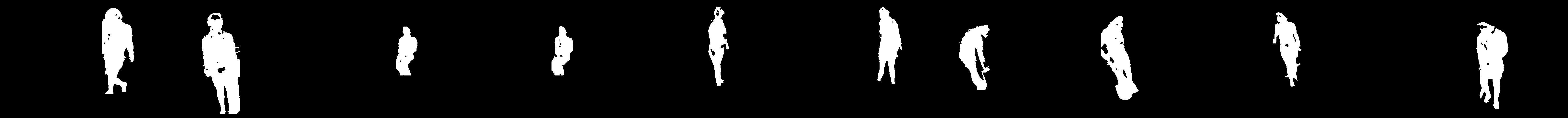}}\\
\vspace{-4mm}
\addtocounter{subfigure}{-1}
\subfigure{\includegraphics[width=\textwidth]{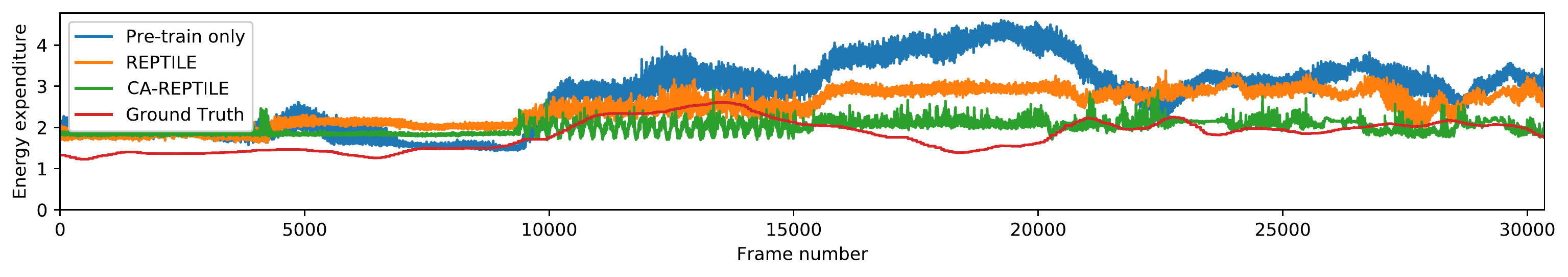}}\vspace{-10pt}
\\
\subfigure{\includegraphics[width=\textwidth, trim={-250 0 -40 0}]{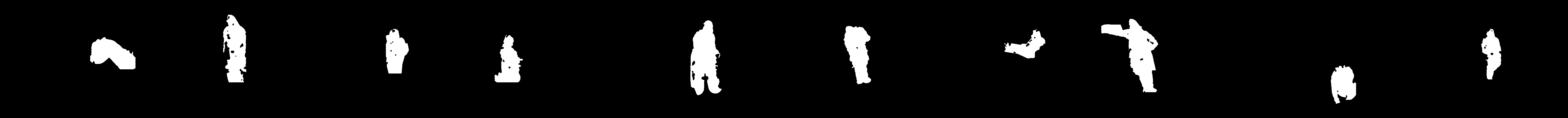}}\\
\vspace{-4mm}
\addtocounter{subfigure}{-1}
\subfigure{\includegraphics[width=\textwidth]{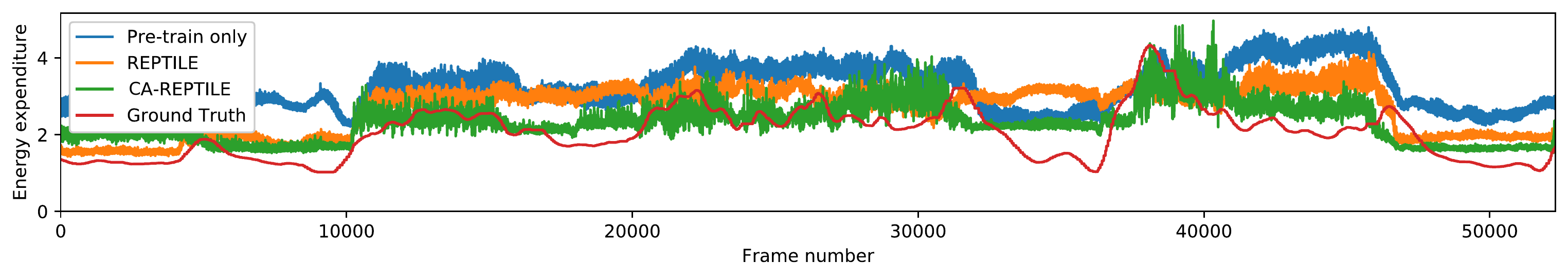}}\vspace{-12pt} 
\caption{Example energy expenditure predictions on two sequences from different participants in the Calorie dataset.}
\label{fig:qual}
\vspace{-10pt}
\end{figure}

Fig.~\ref{fig:qual} shows qualitative silhouette sequences with calorimeter readings as groundtruth, which are to compared to predictions from our method and baselines.
Results demonstrate that the context-agnostic version estimates the ground truth curve better than other methods from participants with low and high energy expenditure variability.

\section{Conclusion}
In this paper, we proposed context-agnostic meta-learning that learns a network initialisation which can be fine-tuned quickly to new few-shot target problems.   An adversarial context network acts on the initialisation in the meta-learning stage, along with task-specialised weights, to learn context-agnostic features capable of adapting to tasks which do not share context with the training set. This overcomes a significant drawback with current few-shot meta-learning approaches, that do not exploit context which is often readily available.  
The framework is evaluated on the Omniglot few-shot character classification dataset and the Mini-ImageNet and CUB few-shot image recognition tasks, where it demonstrates consistent improvements when exploiting context information.  We also evaluate on a few-shot regression problem, for calorie estimation from video, showing significant improvements.  

This is the first work to demonstrate the importance and potential of incorporating context into few-shot methods. We hope this would trigger follow-up works on other problems, methods and contexts.

\noindent \textbf{Data Statement:} Our work uses publicly available datasets. Proposed context-based splits are available at \url{github.com/tobyperrett/context_splits}.

\noindent \textbf{Acknowledgement:} This work was performed under the SPHERE Next Steps Project, funded by EPSRC grant EP/R005273/1.

%
%
\bibliographystyle{splncs}
\bibliography{lib.bib}
\end{document}